\documentclass[11pt,a4paper]{article}
\usepackage[hyperref]{acl2021}
\usepackage{times}
\usepackage{latexsym}

\usepackage{amsmath}

\usepackage{tikz}
\usetikzlibrary{positioning}
\usepackage[framemethod=TikZ]{mdframed}
\mdfsetup{%
 middlelinewidth=0,
 backgroundcolor=gray!15,
 roundcorner=8pt
}

\usepackage{xcolor}  %
\usepackage{graphicx}
\usepackage{url}
\usepackage{multirow}
\usepackage{comment}
\usepackage{booktabs}
\usepackage{colortbl}
\usepackage[subrefformat=parens]{subcaption}

\newcommand{\fscore}{\ensuremath{\mathrm{F}_{0.5}}}

\newcommand{\catS}{\textsc{S}}

\newcommand{\catNP}{\textsc{NP}}
\newcommand{\catVP}{\textsc{VP}}

\newcommand{\catIV}{\textsc{IV}}

\newcommand{\catTV}{\textsc{TV}}
\newcommand{\catQ}{\textsc{Q}}
\newcommand{\catN}{\textsc{N}}

\newcommand{\catAdj}{\textsc{Adj}}
\newcommand{\catAdv}{\textsc{Adv}}

\newcommand{\midspace}{\ensuremath{\ \mid \ }}

\usepackage{microtype}

\aclfinalcopy 
\def\aclpaperid{3614} 

\title{Do Grammatical Error Correction Models Realize \\ Grammatical Generalization?} 

\author{%
Masato Mita\textsuperscript{\textnormal{1, 2}} \textnormal{and}  Hitomi Yanaka\textsuperscript{\textnormal{3, 1}} \\
\textsuperscript{\textnormal{1}}RIKEN AIP,
\textsuperscript{\textnormal{2}}Tohoku University,
\textsuperscript{\textnormal{3}}The University of Tokyo\\
\hypersetup{urlcolor=black} \href{mailto:masato.mita@riken.jp}{\tt masato.mita@riken.jp} \hspace{0.06em} $\vert$ \hspace{0.12em} \href{mailto:hyanaka@is.s.u-tokyo.ac.jp}{\tt hyanaka@is.s.u-tokyo.ac.jp}%
}
\date{}

\begin{document}

\maketitle
\begin{abstract}
There has been an increased interest in data generation approaches to grammatical error correction (GEC) using pseudo data.
However, these approaches suffer from several issues that make them inconvenient for real-world deployment including a demand for large amounts of training data. 
On the other hand, some errors based on grammatical rules may not necessarily require a large amount of data if GEC models can realize grammatical generalization.
This study explores to what extent GEC models generalize grammatical knowledge required for correcting errors.
We introduce an analysis method using synthetic and real GEC datasets with controlled vocabularies to evaluate whether models can generalize to unseen errors.
We found that a current standard Transformer-based GEC model fails to realize grammatical generalization even in simple settings with limited vocabulary and syntax, suggesting that it lacks the generalization ability required to correct errors from provided training examples.

\end{abstract}

\section{Introduction}
\label{sec:intro}
%
Grammatical Error Correction (GEC) is the task of automatically correcting grammatical errors in a text.
GEC's mainstream approach is to consider the task as machine translation (MT) from an ungrammatical text to a grammatical text due to their structural similarity~\citep{brockett-etal-2006-correcting,junczys:2018:NAACL}.
Therefore, many neural encoder-decoder models (EncDec), which are common in MT, have been proposed for GEC, and Transformer-based models have become standard~\cite{grundkiewicz:2018:NAACL,zhao2019improving,kaneko-etal-2020-encoder}.
More recently, there has been an increased interest in data generation approaches to GEC using pseudo data, i.e., improving performance by increasing the amount of training data using pseudo data without making any modifications to the model architecture~\cite{grundkiewicz:2019:bea, kiyono-etal-2019-empirical}.

\begin{figure}[t]
 \centering
  \includegraphics[width=1.0\linewidth]{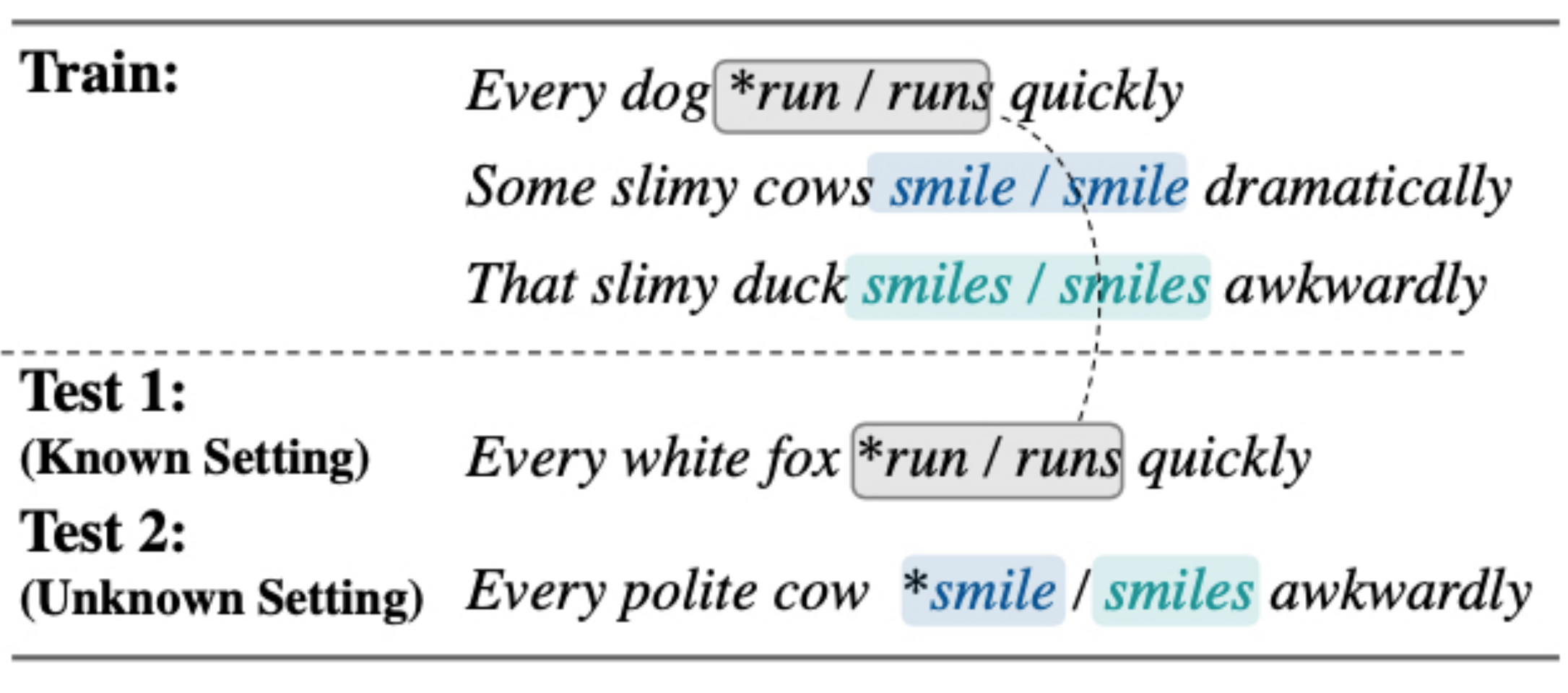}
  \caption{Overview of our proposed method for evaluating the generalization capability of GEC models. In the \emph{Known setting}, the model must correct previously seen patterns. In the \emph{Unknown setting}, the model is presented with an unseen pattern but with familiar vocabulary. We found significantly lower performance in the unknown setting, indicating that the model failed to generalize its grammatical knowledge.}
 \label{fig:overview}
\end{figure}

\begin{table*}[ht]
\centering
\small
\begin{tabular}{lll}
\toprule
Error Type & Synthetic data       &  Real data \\ \midrule
\scriptsize{\textsc{VERB:SVA}}   & \textit{Every white dog \textbf{*run/runs} quickly } & \textit{My mother and father \textbf{*is/are} really an affectionate couple} \\
\scriptsize{\textsc{VERB:FORM}}   & \textit{Some white dogs \textbf{*running/ran} quickly}    & \textit{I am interested in \textbf{*work/working} with you} \\
\scriptsize{\textsc{WO}}       & \textit{\textbf{*White every/Every white} dog ran quickly}  & \textit{I've never seen it \textbf{*before like this/like this before}}  \\
\scriptsize{\textsc{MORPH}}       & \textit{Some white dogs ran \textbf{*quick/quickly} }   & \textit{We have a good \textbf{*relation/relationship} , she is my main friend} \\
\scriptsize{\textsc{NOUN:NUM}}  & \textit{Every \textbf{*dogs/dog} ran } & \textit{You know that I love action \textbf{*film/films} like this} \\ \bottomrule
\end{tabular}
\caption{Examples of automatically constructed synthetic and real data.}\label{tab:datasets}
\end{table*}

However, these approaches suffer from several issues that make them inconvenient for real-world deployment, including a demand for large amounts of training data.
For example, \citet{kiyono-etal-2019-empirical} reported that it was necessary to add about 60 million samples of pseudo-data to improve a standard measure of GEC, \fscore{}, by only two points.
If GEC models can realize grammatical generalization, as humans do not need to memorize individual \textit{error correction patterns} (target terms and its corrections) as long as they have learned grammatical rules, some errors based on grammatical rules (e.g., subject-verb agreement errors) do not necessarily require large amounts of data.

In this study, we explore to what extent GEC models are able to generalize
their grammatical knowledge to correct unseen error correction patterns but with familiar vocabulary.
We propose an analysis method using both synthetic and real datasets, each with controlled vocabularies, to evaluate whether models can generalize to unseen errors (Figure~\ref{fig:overview}).
Experimental results demonstrate that a current standard Transformer-based GEC model does not sufficiently generalize its grammatical knowledge even in simple settings with limited vocabulary and syntax.


\section{Related Work}
\label{sec:related}
Recent studies of probing the syntactic abilities of neural language models have examined whether the models can detect correctness in syntactically challenging tasks such as subject-verb agreement~\cite{linzen-etal-2016-assessing,gulordava-etal-2018-colorless,marvin-linzen-2018-targeted}. 
In contrast, our study focuses on EncDec-based GEC models that not only require a generalized ability to detect errors, but also the ability to \textit{correct} them using information from language modeling and error correction patterns.

In addition, previous studies of probing language models~\citep[i.a.]{gulordava-etal-2018-colorless,marvin-linzen-2018-targeted} often only used synthetic datasets to test models with controlled vocabulary and grammar.
Since GEC models are created to correct data  ``in the wild'', we also use real data in our evaluation and compare performance between data types.

\section{Proposed Method}
\label{sec:proposed_method}

Figure~\ref{fig:overview} shows an overview of the proposed method.
To evaluate the generalization capability of GEC models, we compare the performance when correcting previously seen error correction patterns (\emph{Known setting}) to correcting unseen patterns of the same error type (\emph{Unknown setting}).
Here, an error correction pattern is a pair of terms consisting of a target term (the term with an error that the GEC system needs to correct) and its correction. 
For example, in Figure~\ref{fig:overview},  ``\textit{*run/runs}'' is an error correction pattern that appears in ``\textit{Every dog *run/runs quickly}'' and ``\textit{Every white fox *run/runs quickly}''.
The contexts are different, but both examples need ``\textit{run}'' to be corrected to ``\textit{runs}''.
Here, in the known setting, GEC models must correct other occurrences of ``\textit{run}'' into ``\textit{runs}'' as seen during training, while in the unknown setting, it must also correct unseen error correction patterns such as ``\textit{*smile/smiles}'' that are not appeared in the training data. 
If a model's performance significantly drops in the unknown setting, it indicates a lack of ability to generalize its grammatical knowledge.
%
%
%
%

\begin{table*}[t]
\centering
\small
\begin{tabular}{llrrrrr}
\toprule
       Dataset                &        & \textsc{VERB:SVA} & \textsc{VERB:FORM} & \textsc{WO}    & \textsc{MORPH} & \textsc{NOUN:NUM} \\ \midrule
\multirow{3}{*}{Synthetic data} & Known     & 99.61    & 99.17     & 99.09 & 98.44 &     97.47      \\
                       & Unknown    & 46.05    & 56.93     & 84.00 & 29.35 &       65.55    \\  \rowcolor{gray!20}
                       & $\Delta$    & \bf -53.56    & \bf -42.24     &\bf -15.09 & \bf -69.09 &    \bf -31.92       \\\midrule 
\multirow{3}{*}{Real data} & Known        & 87.84    & 86.36       & 74.89 & 87.77 & 83.75     \\ 
                       & Unknown     & 6.28     & 6.28       & 9.25   & 3.83  & 12.49     \\ \rowcolor{gray!20}
                       & $\Delta$     & \bf -81.56    & \bf -80.08        & \bf -65.64 & \bf -83.94 & \bf -71.26    \\ \bottomrule
\end{tabular}
\caption{Generalization performance for unseen errors. Each number represents an \fscore~score. }\label{tab:main_result}
\end{table*}

We use two types of GEC data: synthetic data and real data (Table~\ref{tab:datasets}).
%
The synthetic data is a fully generated dataset using a set of context-free grammar (CFG) rules and the real data is created by processing existing GEC data.
The purpose of the evaluation using synthetic data is to \textbf{systematically} analyze to what extent the current model achieves the grammatical knowledge generalization required for correcting errors at the architectural level to build the setting with complete control over vocabulary.
While the synthetic dataset offers a fully-controlled environment for precise evaluation, its samples are not representative of data that GEC models are expected to be used for.
To create a more ``natural'' testing environment for comparison, we loosened the strict vocabulary requirement, which is difficult to fulfill with highly varied real data, and recreated the evaluation setup by restructuring existing GEC data.
Note that, due to its softer control, this setting should only be taken as a supplementary comparison for additional insight.

In this study, we investigate standard five error types defined by \citet{bryant:2017:automatic}, which are errors based on grammatical rules: subject-verb agreement errors (VERB:SVA), verb forms errors (VERB:FORM), word order errors (WO), morphological errors (MORPH), and noun number errors (NOUN:NUM).
We created each version of the data as follows.

\paragraph{Synthetic data}
\label{subsec:pseudo_data}
We provide a vocabulary-controlled dataset using CFG inspired by the data generation process in \cite{yanaka-etal-2020-neural}.
More specifically, we design two kinds of generation rules for each of the five error types to be analyzed, one generating grammatical sentences and the other ungrammatical ones\footnote{Appendix ~\ref{appendix:cfg_rule} provides some CFG rules and lexical entries.}.
For example, for VERB:SVA, the rule~\ensuremath{\textsc{S}\rightarrow\textsc{NP}_{pl}\ \textsc{VP}_{sg}} can generate ungrammatical sentences containing ``\textit{*dogs smiles}'', and~\ensuremath{\textsc{S}\rightarrow\textsc{NP}_{sg}\ \textsc{VP}_{sg}} can generate grammatical sentences containing ``\textit{dog smiles}''.
To produce natural sentences, we selected 15 lexical items for nouns, intransitive verbs, transitive verbs, adjectives, and adverbs, respectively.
We can adjust the data size by changing the number of sentences generated by the CFG. 
In this paper, we automatically constructed 50,000 sentence pairs for each error type.

\paragraph{Real data}
\label{subsec:learner_data}

To provide real data, we first perform an automatic annotation of error type labels and error correction patterns on an existing learner dataset using ERRANT~\cite{bryant:2017:automatic}~\footnote{\url{https://github.com/chrisjbryant/errant}}.
Here, we used approximately 2 million sentence pairs as the learner dataset, which is a combination of training and development data distributed by the BEA-2019 Shared Task\footnote{\url{https://www.cl.cam.ac.uk/research/nl/bea 2019st/}}.
Then, we split the data while preserving error types and error correction patterns so that there is one error correction pattern per sentence. 
The unknown setting can be constructed by sorting the entire dataset based on the retained error correction patterns and classifying those with duplicates into training data and those without duplicates into test data.
We constructed the known setting by sampling a small amount of data from training data as test data such that the same error correction patterns are included in both training and test sets. 
Using the above procedure, we obtained 25,889 sentence pairs for \textsc{VERB:SVA}, 41,592 sentence pairs for \textsc{VERB:FORM}, 18,779 sentence pairs for \textsc{WO}, 26,345 sentence pairs for \textsc{MORPH}, and 68,002 sentence pairs for \textsc{NOUN:NUM}.
Compared to the synthetic data, real data has a wide variety of vocabulary and syntax ranging from simple to complex.

\section{Experiments}
\label{sec:exp}

\subsection{Experimental Settings}
We evaluated the grammatical generalization capability of a vanilla Transformer-based EncDec model.
Specifically, we used the \texttt{fairseq} toolkit~\citep{ott2019:arxiv:fairseq} implementation of the ``Transformer (big)'' setting~\cite{Vaswani:17:NIPS}\footnote{See Appendix~\ref{appendix:exp_data_split} and \ref{appendix:hyper-parameter-settings} for details of the datasets and hyperparameters we used, respectively.}, and used the \fscore{} score calculated by ERRANT as the evaluation metric. 
We do not evaluate current state-of-the-art (SOTA) systems for the following two reasons.
First, the top system in BEA2019~\cite{grundkiewicz:2019:bea} and the current SOTA systems~\cite{omelianchuk-etal-2020-gector,kaneko-etal-2020-encoder} use pre-trained models such as pre-trained Masked LMs or use pseudo-data during pre-training. 
A key point of our study is controlling for seen/unseen patterns. 
This becomes difficult with pre-trained models since we cannot know whether a particular pattern is seen during pre-training. 
Second, we believe that evaluating a standard model's architecture, which is commonly used at the core of rapidly evolving SOTA systems, allows for a more accurate analysis by eliminating factors that make analysis more complex, and a more general analysis since our findings can be transferred to most current models, including SOTA systems. 

\subsection{Results}
Table~\ref{tab:main_result} shows the evaluation results.
The evaluation using the synthetic data shows that the model's correction performance drops significantly in the unknown setting compared to the known setting, except for \textsc{WO}.
One reason for the relatively high generalization ability of \textsc{WO} for unseen errors could be its relative simplicity.
Namely, \textsc{WO} can be corrected just by identifying the word's position (Table~\ref{tab:datasets}).
In contrast, other errors need to be corrected while recognizing differences in the surface form of words and dependencies between specific words, which increases the complexity of the correction task.

On the other hand, evaluation using real data show a significant performance drop on all errors, including \textsc{WO}, in the unknown setting, suggesting that generalization is more difficult in more practical settings where the vocabulary and syntax are diverse.

\section{Analysis}
\label{subsec:analysis}
\begin{figure}[t]
 \centering
  \includegraphics[width=1.0\linewidth, height=3cm]{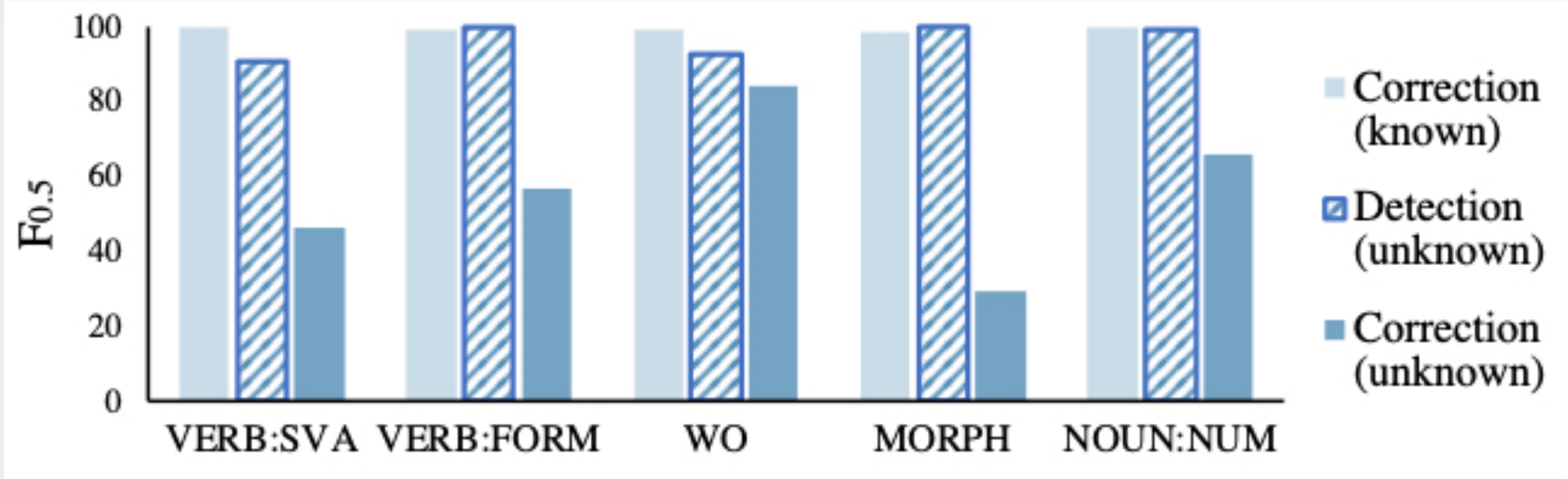}
   \caption{Comparison of detection and correction performance.} \label{fig:enc_dec_all}
\end{figure}

\paragraph{Detection vs. Correction}
To analyze whether the model failed to generalize due to an inability to \emph{detect} errors or an inability to \emph{predict} the correct word, we compare the error detection and correction performance in the unknown setting.
The detection performance is measured by evaluating whether the GEC model makes any edit in the error location.
We evaluated both the detection and the correction performance using ERRANT.
Figure~\ref{fig:enc_dec_all} shows the evaluation result using synthetic data. 
The result shows the model successfully detected all error types, suggesting that the model can generalize its grammatical knowledge at least enough to detect errors, but not enough to predict the correct word.

We can also consider the generalization performance reported in Table~\ref{tab:main_result} as a kind of ablation study: distinguishing, for each error type, how much the language modeling information and the error correction pattern information contribute to improving its correction performance, respectively.
We can assume a model can learn accurate language model information in the unknown setting, but not the error correction patterns.
Therefore, we can see that \textsc{WO}, which has a lower drop in correction performance in the unknown setting compared to the others, can be corrected with language modeling information alone.
This result is consistent with the report~\cite{futrell-levy-2019-rnns} that language models are robust to word order.

\begin{table}[t]
\centering
\begin{tabular}{lrr}
\toprule
          & noiseless & noisy   \\ \midrule
\textsc{VERB:SVA}  &   9.95   & 5.78          \\
\textsc{VERB:FORM} & 12.33     & 5.47        \\
\textsc{WO}        & 7.89 & 9.35     \\
\textsc{MORPH}     &    6.32  & 3.90           \\
\textsc{NOUN:NUM}  &     24.16  &    12.49         \\ \bottomrule
\end{tabular}
\caption{Effect of the complexity of errors in a sentence. Each number represents an \fscore~score.} \label{tab:noiseness2}
\end{table}

\begin{figure}[t]
 \centering
  \includegraphics[width=1.0\linewidth,height=4cm ]{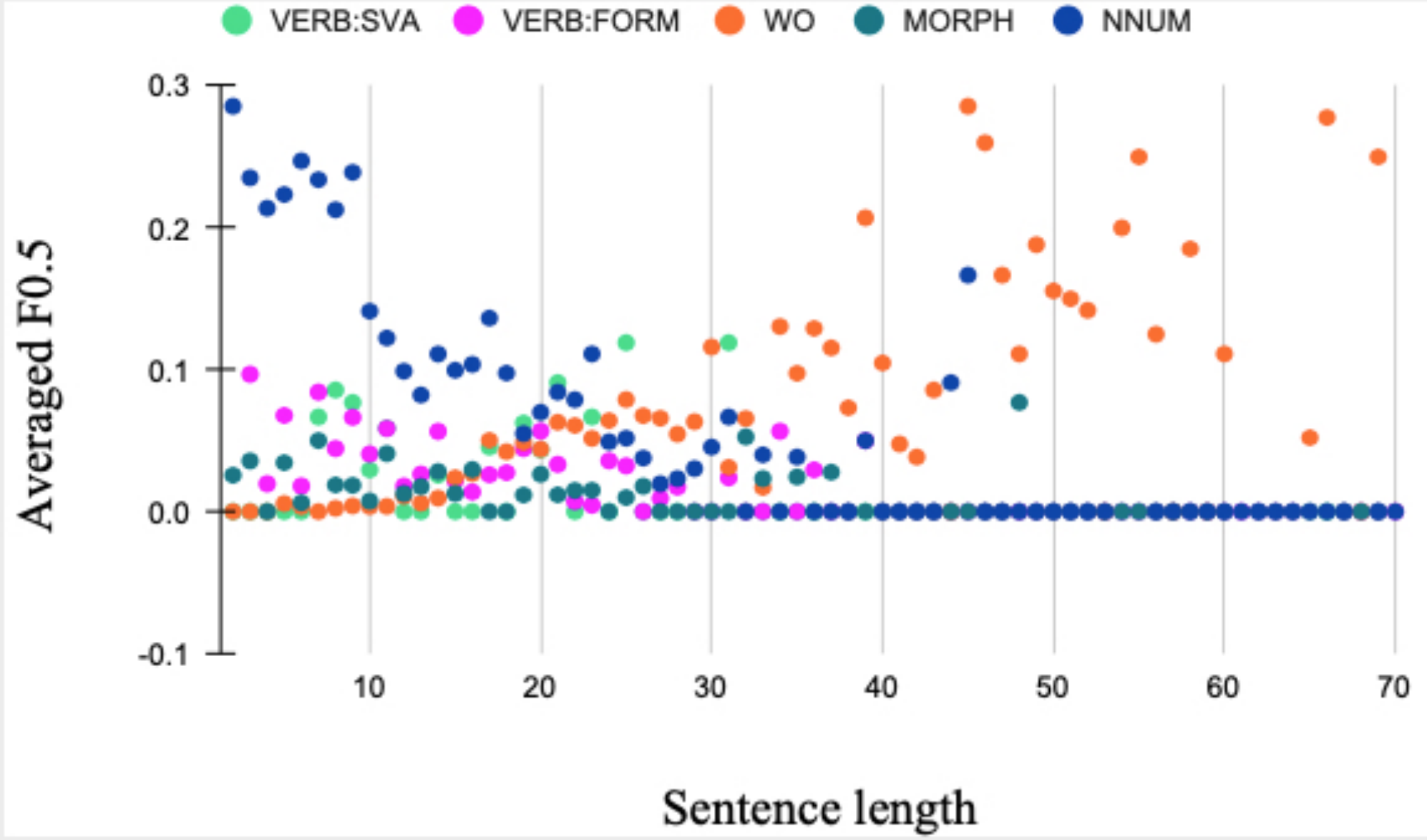}
 \caption{Relationship between sentence length and performance.}
 \label{fig:acc_by_len}
\end{figure}

\begin{table}[t]
\centering
\begin{tabular}{llll}
\toprule
\#seen patterns      & \multicolumn{1}{c}{0} & \multicolumn{1}{c}{1} & \multicolumn{1}{c}{2} \\  \midrule
Precision    & 43.31                 & 47.16                 & 57.65                 \\
Recall    & 47.92                 & 52.52                 & 63.70                 \\
\fscore & 44.16                 & 48.14                 & 58.77         \\ \bottomrule        
\end{tabular}
\caption{Performance change when we expose the model to a few error correction patterns.} \label{tab:few_sample}
\end{table}

\paragraph{Complexity in real data}
To better understand the relationship between complexity and performance, we observed the effect of two contributing factors: error complexity and sentence length. 
Specifically, we compared the performance when the target error is the only error in the sentence (\emph{noiseless}), and when the sentence contains other errors besides the target error (\emph{noisy}).
Table~\ref{tab:noiseness2} shows the effect of the complexity of errors in a sentence.
The results show that the performance of \textsc{WO} is constant with and without noise, while the other errors are affected by noise.
Also, we analyzed the relationship between sentence length and performance (Figure~\ref{fig:acc_by_len}) and confirmed that the difficulty of corrections on \textsc{WO} does not depend on the sentence length.
These results suggest that the reason why the drop in correction performance of WO was relatively low compared to the others, even with real data, is due to its robustness to the complexity of input sentences.

\paragraph{Can a few error correction patterns improve model performance?}
We have found that the current model is vulnerable to unseen errors, but how does its performance change if we expose the model to a few error correction patterns?
Table~\ref{tab:few_sample} shows the performance change when a few error corretion patterns are added to the training data for the pattern ``\textit{*smile/smiles}'' in \textsc{VERB:SVA}.
As test data, we used the test data used in Section~\ref{sec:exp}, excluding sentence pairs other than the target pattern.
From the results, we can see that adding even just one or two samples to the training data can significantly improve the model's performance.
This suggests that when preparing training data for GEC, it is important to sample even one or two seen patterns for each word to improve the performance.

\section{Conclusion}
This study explored to what extent GEC models generalize grammatical knowledge required for correcting errors.
We introduce an analysis method using synthetic and real GEC datasets with controlled vocabularies to evaluate whether models can generalize to unseen errors.
We found that the current standard Transformer-based GEC model can generalize error detection to some extent in a simple synthetic setting, while it cannot generalize correction to a greater extent in both synthetic and real settings, suggesting that it lacks the generalization ability required to correct errors from provided training examples.
Therefore, methods to incorporate grammatical knowledge as rules into the current models can be expected to be necessary to implement a lightweight GEC model requiring less training data, which we plan to investigate in our future work.

\section*{Acknowledgments}
We thank the three anonymous reviewers for their
helpful comments and suggestions.
We are also grateful to Kentaro Inui and Ana Brassard for their insightful comments and suggestions.
This work was partially supported by JSPS KAKENHI Grant
Number JP20K19868.

\bibliographystyle{acl_natbib}
\bibliography{reference}

\newpage
\onecolumn
\appendix

\section{CFG rules used to construct synthetic data}
\label{appendix:cfg_rule}
\begin{table}[h!]
    \centering
    \begin{tabular}{lll} \hline
    \multicolumn{3}{c}{\textbf{Generation rules}} \\
    \catS & $\rightarrow$ & \catNP \ \catVP \\
    \catS$_\text{-SVA}$ & $\rightarrow$ & \catNP$_{sg}$ \ \catVP$_{pl}$ \midspace \catNP$_{pl}$ \ \catVP$_{sg}$\\
    \catVP & $\rightarrow$ & \catIV \midspace \catIV\ \catAdv \midspace \catTV \ \catNP \\
    \catVP$_\text{-FORM}$ & $\rightarrow$ & \catIV$_{ing}$ \midspace \catIV$_{ing}$\ \catAdv \midspace \catTV$_{ing}$ \ \catNP\\
    \catVP$_\text{-MORPH}$ & $\rightarrow$ & \catIV\ \catAdj \\
    \catNP & $\rightarrow$ & \catQ \ \catN \midspace \catQ \ \catAdj \ \catN \\
    \catNP$_\text{-WO}$ & $\rightarrow$ & \catAdj \ \catQ \ \catN\\
    \catNP$_\text{-NUM}$ & $\rightarrow$ & \catQ$_{sg}$ \ \catN$_{pl}$ \midspace \catQ$_{pl}$ \ \catN$_{sg}$ \midspace \catQ$_{sg}$ \ \catAdj \ \catN$_{pl}$ \midspace \catQ$_{pl}$ \ \catAdj \ \catN$_{sg}$\\ \hline
    \multicolumn{3}{c}{\textbf{Lexical items}} \\ 
    \catQ & $\rightarrow$&\{\textit{a, every, no, some, many}\} \\
    \catN &$\rightarrow$&\{\textit{dog, rabbit, cat, bear, tiger}\}\\
    \catIV &$\rightarrow$&\{\textit{run, walk, come, dance, leave}\}\\
    \catTV &$\rightarrow$&\{\textit{kicked, hit, cleaned, touched, accepted}\}\\
    \catAdj &$\rightarrow$&\{\textit{white, gray, big, small, large, old}\}\\ 
    \catAdv &$\rightarrow$&\{\textit{quickly, slowly, gracefully, seriously, happily}\}\\ 
    \hline
    \end{tabular}
    \caption{Examples of CFG rules used for synthetic data construction. The generation rules with errors for each error type are shown by \catVP$_\text{-error type}$ for instance.}
    \label{tab:cfg}
\end{table}

\section{Details of the datasets used in the experiments}
\label{appendix:exp_data_split}
\begin{table}[h!]
\scriptsize{}
\centering
\begin{tabular}{llllll}
\toprule
   & \multicolumn{1}{c}{\textsc{VERB:SVA}} & \multicolumn{1}{c}{\textsc{VERB:FORM}} & \multicolumn{1}{c}{\textsc{WO}} & \multicolumn{1}{c}{\textsc{MORPH}} & \multicolumn{1}{c}{\textsc{NOUN:NUM}} \\ \midrule
Known & 50,000 / 2,000 / 18,562             & 50,000 / 2,000 / 10,125              & 50,000 / 2,000 / 8,438        & 50,000 / 2,000 / 10,125          & 50,000 / 2,000 / 8,438              \\
Unknown & 50,000 / 2,000 / 13,749             & 50,000 / 2,000 /  7,500               & 50,000 / 2,000 / 6,250        & 50,000 / 2,000 / 7,500           & 50,000 / 2,000 / 6,250            \\ 
 \bottomrule
\end{tabular}
\caption{Details of the data split in the synthetic data setting (training/development/test).}
\end{table}

\begin{table}[h!]
\scriptsize{}
\centering
\begin{tabular}{llllll}
\toprule
   & \multicolumn{1}{c}{\textsc{VERB:SVA}} & \multicolumn{1}{c}{\textsc{VERB:FORM}} & \multicolumn{1}{c}{\textsc{WO}} & \multicolumn{1}{c}{\textsc{MORPH}} & \multicolumn{1}{c}{\textsc{NOUN:NUM}} \\ \midrule
Known & 23,889 / 2,000 / 2,000             & 39,592 / 2,000 / 2,000              & 16,779 / 2,000 / 2,000        & 24,345 / 2,000 / 2,000         & 66,002 / 2,000 / 2,000              \\
Unknown & 23,889 / 2,000 / 633             & 34905 / 2,000 /  2000               & 16,779 / 2,000 / 9,199        & 24,345 / 2,000 / 5227           & 66,002 / 2,000 / 3,111            \\ 
 \bottomrule
\end{tabular}
\caption{Details of the data split in the real data setting (training/development/test).}
\end{table}

\newpage
\section{Hyper-parameter settings}
\label{appendix:hyper-parameter-settings}

\begin{table}[h!]
\centering
\begin{tabular}{@{}lp{100mm}@{}}
\toprule
 Configurations & Values \\ \midrule
 Model Architecture & Transformer~\citep{Vaswani:17:NIPS} \\
 Optimizer& Adam~\citep{kingma:2015:ICLR}\\
 Learning Rate Schedule & Same as described in Section 5.3 of \citep{Vaswani:17:NIPS} \\
 Number of Epochs & 30 for synthetic data and 150 for real data\\
 Dropout& 0.3\\
 Stopping Criterion & Train model for 30 epochs (synthetic data) and 150 epochs (real data). \\ 
 Gradient Clipping& 1.0\\
 Loss Function& Label smoothed cross entropy~\citep{szegedy:2016:rethinking} \\
 Beam Search& Beam size 5 with length normalization\\
 \bottomrule
\end{tabular}
\caption{Detailed hyper-parameters used for the base GEC model.}
\end{table}

\end{document}